\pdfoutput=1

\documentclass[sigconf]{acmart}
\AtBeginDocument{%
  }

\usepackage{multirow}
\usepackage[normalem]{ulem}
\useunder{\uline}{\ul}{}
\usepackage{booktabs}
\usepackage{array}
\usepackage{makecell}

\renewcommand{\arraystretch}{0.85}
\newcolumntype{M}[1]{>{\centering\arraybackslash}m{#1}}

\begin{document}

\title{Learning Robust Heterogeneous Graph Representations via Contrastive-Reconstruction under Sparse Semantics}

\author{Di Lin}
    \affiliation{%
    \institution{College of Software, Jilin University}
    \city{Changchun}
    \state{Jilin}
    \country{China}
}

\author{Wanjing Ren}
    \affiliation{%
    \institution{College of Software, Jilin University}
    \city{Changchun}
    \state{Jilin}
    \country{China}
}

\author{Xuanbin Li}
    \affiliation{%
    \institution{College of Software, Jilin University}
    \city{Changchun}
    \state{Jilin}
    \country{China}
}

\author{Rui Zhang}
    \affiliation{%
    \institution{College of Computer Science and Technology, Jilin University}
    \city{Changchun}
    \state{Jilin}
    \country{China}
}

\renewcommand{\shortauthors}{Trovato et al.}

\begin{abstract}

In graph self-supervised learning, masked autoencoders (MAE) and contrastive learning (CL) are two prominent paradigms. MAE focuses on reconstructing masked elements, while CL maximizes similarity between augmented graph views. Recent studies highlight their complementarity: MAE excels at local feature capture, and CL at global information extraction. Hybrid frameworks for homogeneous graphs have been proposed, but face challenges in designing shared encoders to meet the semantic requirements of both tasks. In semantically sparse scenarios, CL struggles with view construction, and gradient imbalance between positive and negative samples persists.
This paper introduces HetCRF, a novel dual-channel self-supervised learning framework for heterogeneous graphs. HetCRF uses a two-stage aggregation strategy to adapt embedding semantics, making it compatible with both MAE and CL. To address semantic sparsity, it enhances encoder output for view construction instead of relying on raw features, improving efficiency. Two positive sample augmentation strategies are also proposed to balance gradient contributions.
Node classification experiments on four real-world heterogeneous graph datasets demonstrate that HetCRF outperforms state-of-the-art baselines. On datasets with missing node features, such as Aminer and Freebase, at a 40\% label rate in node classification, HetCRF improves the Macro-F1 score by 2.75\% and 2.2\% respectively compared to the second-best baseline, validating its effectiveness and superiority.
  
\end{abstract}





\maketitle

\section{Introduction}
Heterogeneous Information Networks (HINs) \cite{sun2012mining} are widely used in real-world applications, including academic networks, knowledge graphs, and social networks \cite{wang2017knowledge}. Unlike homogeneous graphs that contain only a single node type and edge type, HINs comprise multiple types of nodes and edges \cite{zheng2022graph}. This heterogeneous structure enables HINs to more accurately model complex real-world relationships and encode richer semantic information\cite{shi2016survey}. To enrich information in HINs, nodes are often associated with labels. Despite the superior performance of graph neural networks (GNNs\cite{scarselli2008graph}) in classifying HIN nodes, most existing HGNN\cite{zhang2019heterogeneous} models rely heavily on large-scale labeled data, which is costly to obtain. To address this, self-supervised learning has been applied to HINs.

Currently, generative learning and contrastive learning are two primary self-supervised learning paradigms for heterogeneous graph representation learning \cite{liu2021self}.  
Generative learning captures complex relationships by reconstructing graph structures and generates node representations from neighborhood information \cite{zheng2022graph}, but it relies on shallow graph neural networks (GNNs), which are limited to capturing $k$-hop neighborhood information and unable to model the global structure of graphs \cite{wang2024rethinking}.  
Contrastive learning constructs positive and negative sample pairs to capture global topological patterns by maximizing similarity between positive pairs and minimizing similarity between negative pairs \cite{chen2020simple,zhu2021graph}, but it requires effective data augmentation strategies to construct high-quality views \cite{zhang2024graph} and lacks the ability to reconstruct node features and edges, resulting in weak local information capture \cite{jiang2024localgcl}.  
The complementary advantages and disadvantages of these two methods have driven the research on hybrid frameworks integrating generative and contrastive learning methods.

\begin{figure}[htbp]
    \centering
    \includegraphics[width=0.45\textwidth]{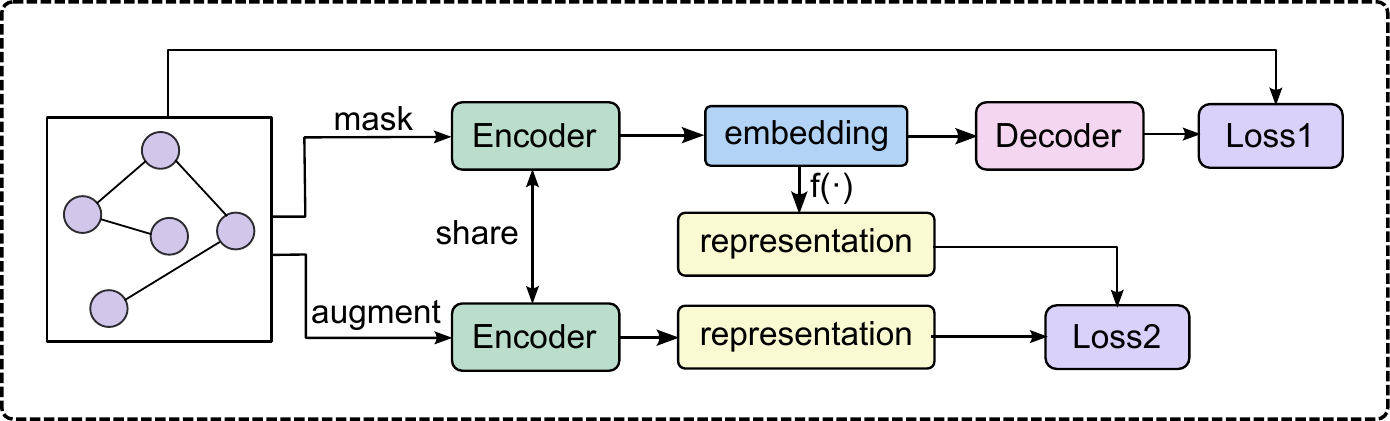}
    \caption{Existing GCMAE model architecture}
    \Description{}
    \label{fig:GCMAE}
\end{figure}

Existing hybrid frameworks typically divide the model architecture into a generative channel and a contrastive channel. Meanwhile, GCMAE\cite{wang2024generative}(The model architecture is shown in Figure \ref{fig:GCMAE}.) points out the drawbacks of using independent encoders for two channels and proposes a mainstream hybrid framework where the two channels share an encoder. However, this framework still has certain limitations. Specifically, generative learning reconstructs graph structures or feature representations by aggregating information from neighboring nodes\cite{yang2025nodereg}. This process tends to cause over-smoothing, making it more suitable for shallower encoder structures to preserve semantic details. In contrast, contrastive learning relies on specially designed objective functions to explicitly suppress embedding homogeneity. This allows for deeper GNN architectures that can effectively aggregate high-order neighborhood information and enhance the modeling of global structures. Due to the significant differences in the focus and semantic representation requirements between these two learning paradigms, a shared encoder struggles to satisfy both, thereby constraining overall performance. Furthermore, when contrastive learning is applied to heterogeneous information networks (HINs)\cite{liu2023hierarchical}, semantic information is sometimes sparse due to issues like missing node features. View augmentations can further disrupt this already sparse semantic structure, making it difficult to generate high-quality contrastive views that preserve semantic invariance\cite{arantes2022learning}. Additionally, contrastive learning loss functions often suffer from imbalanced contributions between positive and negative samples, where positive samples disproportionately influence model training\cite{wang2023positive}. This leads the model to focus excessively on local similarity while neglecting effective capture and understanding of global structure, thus degrading the global semantic modeling capability of contrastive methods. Therefore, the key challenges for self-supervised learning on HINs are summarized as follows:

\begin{itemize}

\item There is a mismatch in the embedding representation requirements between the generative and contrastive channels, making it difficult for a shared encoder to simultaneously satisfy their differing demands for semantic expression.

\item In heterogeneous information networks (HINs), especially under semantic sparsity \cite{you2020does}(e.g., missing original features), it is challenging to construct contrastive views that are both informative and discriminative \cite{chen2020simple}.

\item In contrastive learning, there is an imbalance in gradient contributions between positive and negative samples, which causes the model to overly focus on the features of individual nodes. This can lead to a degenerate state, thereby impairing the model’s ability to effectively capture global structural information.

\end{itemize}

To address these issues, this paper proposes a self-supervised learning framework named HetCRF. Built upon the existing architecture where the generative and contrastive channels share an encoder, a Graph Convolutional Network (GCN\cite{kipf2016semi}) is added to the contrastive channel for secondary aggregation. This aims to obtain richer semantics required by the contrastive channel, resolving the challenge that the encoder struggles to simultaneously meet the distinct semantic expression demands of the two channels.

Unlike existing methods that perform augmentation on raw data, HetCRF constructs contrastive views by augmenting embeddings derived from the encoder to extract deep invariances within the views \cite{you2020does}. In scenarios with sparse semantics (e.g., missing original node features), existing methods struggle to preserve semantic invariance when constructing contrastive views via data augmentation. HetCRF addresses this by augmenting embeddings with richer semantic information. Meanwhile, based on the characteristics of GCN, we construct augmented views from structural attention and semantic attention\cite{qin2021relation}, aiming to align the two attentions to further improve the model's information extraction ability.  

Finally, HetCRF identifies the problem of gradient imbalance between positive and negative samples through theoretical analysis and adopts two positive sample augmentation strategies to balance the gradient difference between positive and negative samples, alleviating gradient dilution from negative samples. This strategy enhances the ability of contrastive learning to extract global information.

The contributions of this paper are summarized as follows:

\begin{itemize}

\item This paper proposes a novel self-supervised hybrid framework HetCRF for heterogeneous graphs, addressing the challenge that encoders cannot simultaneously meet the semantic requirements of two tasks. 

\item Compared with existing methods, we design augmented views on encoder embeddings to address the issue of semantic sparsity. Additionally, we develop novel augmentation strategies based on the characteristics of GCN to preserve more semantic invariance.

\item Two positive sample augmentation strategies are employed to balance the gradient discrepancy between positive and negative samples, enhancing the capability of contrastive learning to extract global information.

\item Extensive experiments show that HetCRF significantly outperforms existing generative, contrastive, and hybrid framework baselines on multiple heterogeneous information network (HIN) datasets. In scenarios with semantic sparsity, on the AMiner and Freebase datasets with a label rate of 40\%, the Macro-F1 score of HetCRF increased by 2.75\% and 2.2\%, respectively, compared to suboptimal baselines. These results strongly validate its superior performance in semantically sparse scenarios.

\end{itemize}

\section{Related Work}

\subsection{Graph Contrastive Learning}

Contrastive methods aim to enhance representation discrimination by pulling together positive sample pairs and pushing apart negative pairs in the embedding space\cite{yao2022pcl}. These methods have demonstrated strong performance in graph neural networks. Representative recent works on heterogeneous graphs include MEOW\cite{yu2024heterogeneous}, ASHGCL\cite{jiang2025incorporating}, and HGMS-C\cite{wang2025homophily}. MEOW introduces a hard negative sampling strategy by applying negative sample weighting. ASHGCL captures invariance among multiple semantics by constructing three views: low-order relations, high-order relations, and feature similarity. HGMS-C strengthens semantic invariance across views through a selective edge-dropping strategy. Overall, the core idea of contrastive learning in training encoders lies in extracting semantic invariance across different views\cite{lin2022cclsl}. By modeling semantic correlations among views, the encoder can capture global information. However, in semantically sparse scenarios (e.g., missing original node features), data augmentation often disrupts the existing sparse semantics, making it difficult to maintain semantic invariance. This problem remains largely unresolved.

\subsection{Graph Generative Learning}

Generative graph self-supervised learning focuses on reconstructing graph structures or node features, guiding the model to learn generalized representations by recovering masked information\cite{he2022masked}. For example, GraphMAE\cite {hou2022graphmae} introduces a masked autoencoder framework to reconstruct masked node features. Building upon this, HGMAE\cite {tian2023heterogeneous} proposes a dynamic masking scheme that includes meta-path masking and adaptive feature masking. It also introduces three reconstruction tasks, including edge structure recovery, feature restoration, and position prediction, to enhance semantic modeling in heterogeneous graphs. Although generative methods excel at capturing structural patterns, the learned representations are limited to local neighborhood information and fail to capture global graph characteristics due to the constraints of shallow encoders\cite{wang2024generative}.

\subsection{Hybrid Learning Framework}

Due to the differences in information capture levels between generative and contrastive learning paradigms, hybrid frameworks integrating both have been proposed in recent years. In homogeneous graph scenarios, typical approaches like CGMP-GL\cite{tang2025generative} leverage independent encoders in generative and contrastive channels to learn global information separately, then support downstream tasks through embedding aggregation; GCMAE\cite{wang2024generative} adopts a single encoder sharing mechanism to achieve collaborative capture of local and global information, generating more complete node representations.  
In the field of heterogeneous graphs, the recently proposed GC-HGNN\cite{wang2025generative} framework introduces the generative-contrastive hybrid architecture into heterogeneous data scenarios for the first time by designing a heterogeneous information aggregation mechanism.

\section{Method}

In this section, we introduce the proposed HetCRF framework. A formal definition of heterogeneous graphs is provided as follows:

For a heterogeneous graph \( \mathcal{G}=\left(\mathcal{V}, \mathcal{E}, A, \mathcal{X}, {\mathcal{T}_{v}}, \mathcal{T}_{\varepsilon}, \Phi\right) \), \( \mathcal{V} \) represents the set of nodes, \( \mathcal{E} \) represents the set of edges,  \( A \) represents the adjacency matrix, \( \mathcal{X} \) represents the feature matrix, \( \mathcal{T}_{v} \) represents the types of node $v$, \( \mathcal{T_{\varepsilon}} \) represents the types of edge $e$ and \( \Phi \) represents the set of meta-paths.  For example, \( \mathcal{T}_{\nu_1} \xrightarrow{\mathcal{T}_{\epsilon_1}} \mathcal{T}_{\nu_2} \xrightarrow{\mathcal{T}_{\epsilon_2}} \cdots \xrightarrow{\mathcal{T}_{\epsilon_l}} \mathcal{T}_{\nu_{l+1}} \) is a meta-path of length \( l \).  \(A^{\phi} \) ,\(  {\phi} \in {\Phi} \)  is the adjacency matrix based on the meta-path  \( \phi\).

\begin{figure*}[htbp]
    \centering
    \includegraphics[width=\textwidth]{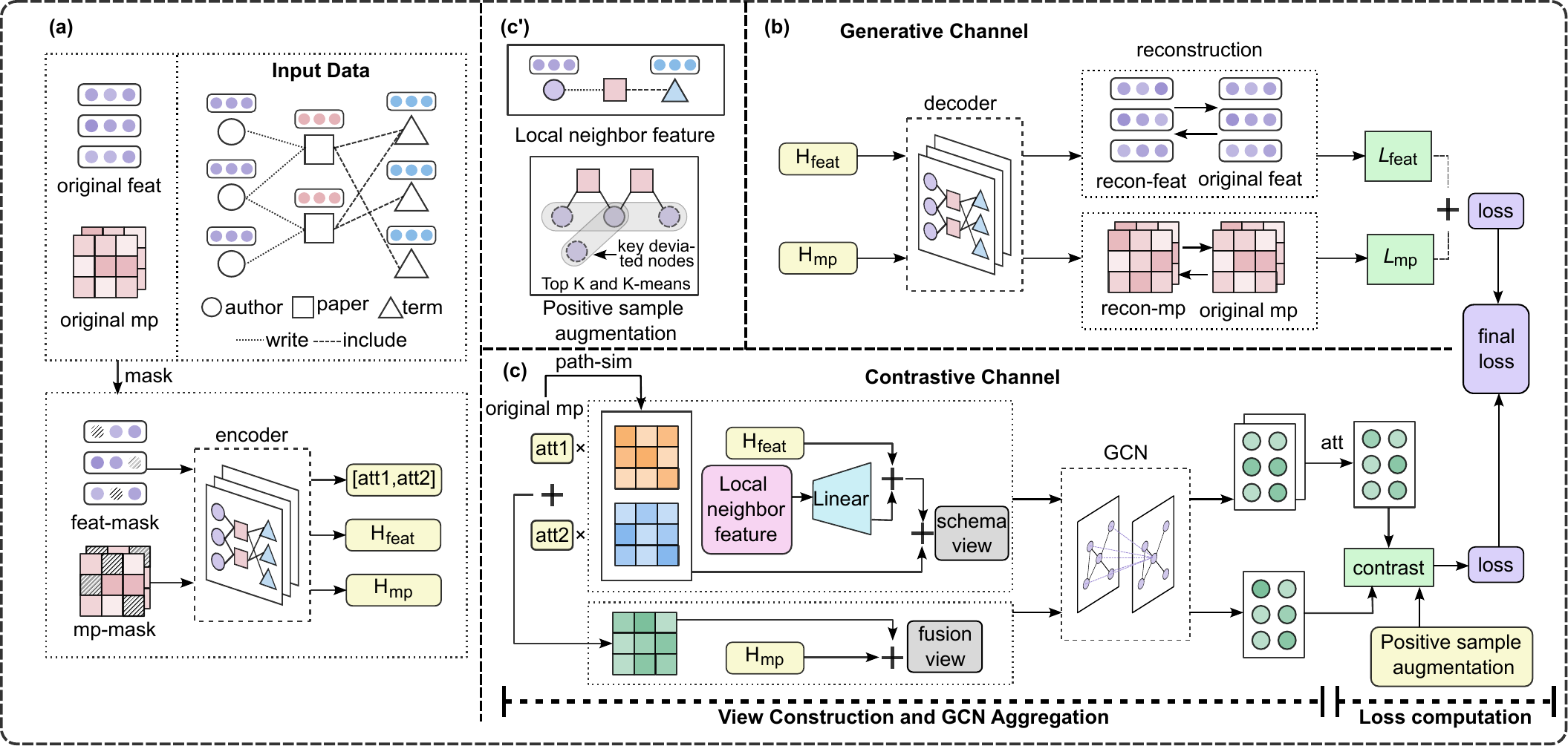}
    \caption{The overall HetCRF framework comprises two channels: the generative channel (b) and the contrastive channel (c). Box (a) illustrates the input data and the encoding process applied to the input data. First, masking strategies are applied to the feature matrix and adjacency matrix, with masked inputs fed into encoders to generate two node embeddings, \(H_{\text{feat}}\) and \(H_{\text{mp}}\). The generative channel uses \(H_{\text{feat}}\) and \(H_{\text{mp}}\) for reconstruction tasks. The contrastive channel constructs contrastive views from \(H_{\text{feat}}\) and \(H_{\text{mp}}\), which are then input into a GCN for second-order aggregation and subjected to contrastive optimization based on augmented positive samples. Here, (c') denotes the illustration of heterogeneous neighbor node features and positive sample augmentation, which are utilized in (c).}
    \label{fig:model}
    \Description{}
\end{figure*}

The overall architecture of HetCRF is illustrated in \textbf{Figure \ref{fig:model}}. The model consists of two main components: a generative channel and a contrastive channel. The generative channel focuses on the information at the node level\cite{li2025p2gcn}, while the contrastive channel targets the global graph-level information. These two components will be described in detail in the following subsections.

\subsection{Generative Channel}

To extract richer node-level features, HetCRF employs two reconstruction strategies: feature reconstruction and meta-path reconstruction. Feature reconstruction masks node features to force the model to recover them using neighborhood information, enhancing local aggregation. Meta-path reconstruction masks meta-paths to help the model learn local structural features of nodes\cite{fu2020magnn}.

\paragraph{\textbf{Feature Reconstruction.}}

At the feature level, a masking strategy similar to GraphMAE\cite{hou2022graphmae} is used. A mask rate $\rho \in (0,1)$ is set, and $\rho \lvert \mathcal{V} \rvert$ nodes are randomly selected.
$ \tilde{\mathcal{V}}$ is the set of masked nodes. Their feature vectors are replaced with a learnable $[MASK]$ token. Then, the masked feature matrix $\tilde{\mathcal{X}}$ and adjacency matrix set ${A^{\Phi}}$ are fed into the HAN encoder $g_{E}$ to obtain node representations $H_{feat}$: $H_{feat} = g_{E}({A}^{\Phi}, \tilde{\mathcal{X}})$.

To enhance the expressiveness of the decoder, we apply the same masking operation to $H_{feat}$ again, yielding $\widetilde{H}_{feat}$. Next, the HAN decoder $g_D$ receives ${A}$ and $\widetilde{H}_{feat}$ to reconstruct node features $Z$: $Z = g_{D}({A}^{\Phi},\widetilde{H}_{feat})$.

The reconstruction loss is calculated between $Z$ and the original feature matrix \( \mathcal{X} \) using scaled cosine error, where $\gamma_{1}$ is the scaling factor:

\begin{equation}
\mathcal{L}_{\text{feat}} = \frac{1}{|\widetilde{\mathcal{V}}|} \sum_{v \in \widetilde{\mathcal{V}}} \left(1 - \frac{\mathcal{X}_v \cdot Z_v}{|\mathcal{X}_v| \times |Z_v|}\right)^{\gamma_1}
\end{equation}

\paragraph{\textbf{Meta-Path Reconstruction.}}

At the meta-path level, a random masking strategy is applied for directed graphs. Concretely, for each \(A^{\phi}\), we create a binary mask following the Bernoulli distribution \(M_{A}^{\phi} \sim \text{Bernoulli}(p_e)\), where \(p_e < 1\) is the edge masking rate. Then, we leverage the \(M_{A}^{\phi}\) to obtain the masked meta-path based adjacency matrix \(\widetilde{A}^{\phi}=M_{A}^{\phi}\cdot A^{\phi}\).

The set of masked adjacency matrices is denoted as $\widetilde{A}^{\Phi}$. This matrix, along with the feature matrix $\mathcal{X}$, is fed into the HAN encoder to obtain $H_{mp}$: $H_{mp} = g_{E}(\widetilde{A}^{\Phi}, \mathcal{X})$

Then, $H_{mp}$ and $\widetilde{A}_{\Phi}$ are input to the decoder to obtain the reconstructed embedding $Z_{mp}^{\Phi}$: $Z_{mp}^{\Phi} = g_{D}(\widetilde{A}^{\Phi}, H_{mp})$

The reconstructed adjacency matrix $A^{\prime}$ is generated by computing the inner product of node embeddings followed by a Sigmoid activation: $A^{\prime} = \sigma((Z_{mp}^{\Phi})^T \cdot Z_{mp}^{\Phi})$

Here, $\sigma$ represents the Sigmoid function. The reconstruction loss is computed using scaled cosine error, where $\gamma_{2}$ is a scaling factor:

\begin{equation}
\mathcal{L}^\phi = \frac{1}{|A^\phi|} \sum_{v \in \mathcal{V}} \left(1 - \frac{A_v^\phi \cdot A_v^{\prime}}{|A_v^\phi| \times |A_v^{\prime}|}\right)^{\gamma_2} , \phi \in \Phi
\end{equation}

To integrate the loss across all meta-paths $\phi$, a semantic-level attention vector $q$ is introduced to learn the importance score $s^\phi$ of each meta-path. These scores are normalized using a softmax function to produce weights $\alpha^\phi$:

\begin{equation}
s^\phi = q^T \cdot \tanh(W \cdot H^\phi + b), \quad \alpha^\phi = \frac{\exp(s^\phi)}{\sum_{\varphi \in \Phi} \exp(s^\varphi)}
\label{eq:3}
\end{equation}

Here, $H^\phi$ is the embedding of the node obtained from the meta-path $\phi$. $W$ is a weight matrix and $b$ is a bias vector. The final meta-path reconstruction loss $\mathcal{L}_{\text{mp}}$ is obtained by combining individual losses using the learned attention weights:

\begin{equation}
\mathcal{L}_{\text{mp}} = \sum_{\phi \in \Phi} \alpha^\phi \cdot \mathcal{L}^\phi
\end{equation}

\subsection{Contrastive Channel}

Based on the encoded features $H_{feat}$ and $H_{mp}$, additional noise is introduced to encourage the model to capture more essential invariances of the graph\cite{gianinazzi2021learning}.

\subsubsection{View Construction}

\paragraph{\textbf{Schema View.}}

Graph schema-level structural information is introduced to enhance representations\cite{gottron2015analysis}. Specifically, based on $H_{feat}$, information from heterogeneous neighbors (e.g., A-P, A-P-T)\cite{zhang2020hop} is aggregated to obtain enriched node representations $H_{\text{agg}}$:

\begin{equation}
H_\text{agg}^{i} = \sigma \left( h_i + \sum_{ \mathcal{T}_{j}  \neq  \mathcal{T}_{i}} \sum_{v \in N_i^{\mathcal{T}_j}} W_{\mathcal{T}_{j}} \cdot h_v \right)
\end{equation}

Here \(h_i\) denotes the representation of node $i$ in $H_{feat}$. $j$ refers to the type of heterogeneous neighboring nodes of $i$. $N_i^{j}$ refers to the set of neighboring nodes of type $j$ of node $i$. $ W_{j}$ denotes the transformation matrix mapping nodes of type $j$ from their Hermitian space to that of the target node. 

The adjacency matrix corresponding to each meta-path is first extracted, forming a collection denoted as $A^{\Phi}$. For each adjacency matrix $A^{\phi} \in A^{\Phi}$, path similarity is applied to select the Top-K neighbors\cite{sun2011pathsim}, filtering out unimportant edges and reducing the size of the fused matrix.

Given a meta-path $\phi$, the similarity between two nodes $i$ and $j$ of the same type is computed as:

\begin{equation}
PS(i, j) = \frac{2 \times \left| { p_{i \rightsquigarrow j} \mid p_{i \rightsquigarrow j} \vdash \phi } \right|}{\left| { p_{i \rightsquigarrow i} \mid p_{i \rightsquigarrow i} \vdash \phi } \right| + \left| { p_{j \rightsquigarrow j} \mid p_{j \rightsquigarrow j} \vdash \phi } \right|}
\end{equation}

Here, $p_{i \rightsquigarrow j}$ denotes a path instance between $i$ and $j$ under meta-path $\phi$. Based on the similarity score, the top $K$ most similar neighbors are selected for each node. After filtering, the adjacency matrix induced by meta-path $\phi$ is denoted as $A_{sim}^\Phi$. Together with the aggregated representations $H_{\text{agg}}$ , this forms the schema view, which jointly captures structural and semantic relationships.

Subsequently, $H_{\text{agg}}$ and the normalized adjacency matrices $\widetilde{A}_{sim}^\phi = D^{-\frac{1}{2}} \cdot A_{sim}^\phi \cdot D^{-\frac{1}{2}}$ are fed into a GCN with the propagation rule $f(x) = \delta(\hat{A}g(x)\cdot W)$, where $\hat{A}$ is the normalized adjacency matrix, $W$ is the weight matrix, and $\sigma$ denotes the activation function. This message passing process produces the final representations $z^\phi$. Here, $D$ denotes the degree diagonal matrix where each element $D_{ii} = \sum_{j=1}^{|\mathcal{V}|} {A_{sim}^{\phi}}_{ij}$ represents the sum of edge weights for node $i$.

Since \(A_{sim}^{\Phi}\) represents a set of meta-path adjacency matrices, for each \(A_{sim}^\phi \in A_{sim}^{\Phi}\), a corresponding embedding \(z^\phi\) can be derived by feeding it into GCN. These embeddings are then aggregated using an attention mechanism:

\begin{equation}
z^\phi = \delta(\widetilde{A}_{sim}^\phi \cdot \text{H}_{agg} \cdot W), \phi \in \Phi
\end{equation}

\begin{equation}
w^{\phi} = \frac{1}{|\mathcal{V}|} \sum_{i \in \mathcal{V}} \mathbf{a}^T \cdot \tanh(W_{\text{att}} \cdot z_i^{\phi} + b_{\text{att}})
\end{equation}

\begin{equation}
\beta^{\phi} = \frac{\exp(w^{\phi})}{\sum_{j=1}^{|Z_i|} \exp(w^j)} \  , \ 
Z_{\text{Schema}} = \sum_{\phi \in \Phi} \beta^\phi \cdot z^\phi\end{equation}

Here, attention weights $\beta^{\phi}$ are used to measure the importance of embeddings from different meta-paths. $W_{\text{att}} \in \mathbf{R}^{d \times d}$ is the weight matrix, and $b_{\text{att}}$ is the bias vector. \(Z_{\text{Schema}}\) is the embedding aggregated via semantic attention.

\paragraph{\textbf{Fusion View.}}

The fused adjacency matrix $A_{fusion}$ is computed by aggregating the path similarity-based matrices $\widetilde{A}_{sim}^{\Phi}$ with attention-weighted integration:

\begin{equation}
A_{\text{fusion}} = \sum_{\phi \in \Phi} \alpha^\phi \cdot \widetilde{A}_{sim}^\phi
\end{equation}

The attention weights are shared with the generative channel and are denoted as $\alpha^{\phi}$. The resulting fused adjacency matrix $A_{\text{fusion}}$ is combined with $H_{mp}$ to form the fusion view. 

$\alpha^{\phi}$ is the attention coefficient calculated by the generative channel in \textbf{Equation (\ref{eq:3})}. 

Similar to the schema view, $A{\text{fusion}}$ and $H_{mp}$ are passed into a GCN to obtain $Z_{\text{fusion}}$:

\begin{equation}
Z_{\text{fusion}} = \delta \left( \sum_{\phi \in \Phi} \alpha^{\phi} \cdot \widetilde{A}_{sim}^{\phi} \cdot \text{H}_{\text{mp}} \cdot W \right)
\label{eq:1}
\end{equation}

\begin{equation}
Z_{\text{schema}} = \sum_{\phi \in \Phi} \beta^{\phi} \cdot \delta \left(\widetilde{A}_{sim}^{\phi} \cdot H_{\text{agg}} \cdot W \right)
\label{eq:2}
\end{equation}

A comparison between \textbf{Equations (\ref{eq:1})} and \textbf{(\ref{eq:2})} reveals distinct mechanisms in different views. In the fusion view, the weighted aggregation of meta-paths focuses on capturing key structural information. In the schema view, the semantic aggregation of multiple embeddings allows key semantic patterns to be captured. This enables the model to learn deeper semantic invariance across views.

Compared with existing contrastive view settings, HetCRF derive augmented contrastive views from encoder embeddings to address scenarios with sparse original semantics. We employ a Graph Convolutional Network (GCN) for secondary information aggregation to enrich semantic representations required by contrastive channels.

\subsubsection{Positive Sample Augmentation}

\paragraph{\textbf{Theorem 1.}} In the optimization objective of contrastive learning, the gradient contribution of the positive sample of an anchor node is the same as that of all negative samples. Let f($x_i$) represent the learned embedding of node $i$. Given $i$ as an anchor, $j$ as its positive sample, and \(\mathcal{N}_i\)denote the set of negative samples of $i$. Then the theorem shown in the following equation must hold.

\begin{equation}
\sum_{k \in \mathcal{N}_i} \frac{\partial\mathcal{L}_{i}}{\partial f(x_{k})} = - \frac{\partial \mathcal{L}_{i}}{\partial f(x_{j})}
\end{equation}

\paragraph{\textbf{Proof.}}Typically, the dot product is used as a similarity function, and the InfoNCE loss\cite{oord2018representation} can be expressed as:

\begin{equation}
\mathcal{L}_i = - \log \frac{\exp(f(x_i)^T \cdot f(x_j) / \tau)}{\exp(f(x_i)^T f(x_j)/\tau) + \sum_{k \in \mathcal{N}_i} \exp(f(x_i)^T f(x_k) / \tau)}
\end{equation}

First, we can calculate the gradient of a positive sample by taking the derivative.

\begin{equation}
\begin{split}
\frac{\partial \mathcal{L}_i}{\partial f(x_j)} &= \frac{\partial \mathcal{L}_i}{\partial p_{ij}} \cdot \frac{\partial p_{ij}}{\partial (f(x_i) \cdot f(x_j))} \cdot \frac{\partial (f(x_i) \cdot f(x_j))}{\partial f(x_j)} \\
&= -\frac{1}{p_{ij}} \cdot p_{ij} \cdot (1 - p_{ij}) \cdot \frac{1}{\tau} \cdot f(x_i) \\
&= -\frac{1}{\tau} \cdot (1 - p_{ij}) f(x_i)
\end{split}
\end{equation}

Here, \(p_{ij}\) denotes the softmax weight of the similarity between $i$ and $j$, where $z$ is the denominator of the softmax expression. So, $z$ and $p_{ij}$ follow the following definitions:

\begin{equation}
z = \exp(f(x_i)^T f(x_j) / \tau) + \sum_{k \in \mathcal{N}_i} \exp((f(x_i)^T f(x_k) / \tau))
\end{equation}

\begin{equation}
p_{ij} = \frac{\exp(f(x_i)^T f(x_j) / \tau)}{z}
\end{equation}

For a particular negative sample $t$, $t$ $\in \mathcal{N}_i$, the gradient of the negative sample t is:

\begin{equation}
\frac{\partial \mathcal{L}_i}{\partial f(x_t)} = \frac{1}{\tau} \cdot p_{it} \cdot f(x_i)
\end{equation}

The sum of gradients from all negative samples can be expressed as follows:

\begin{equation}
\sum_{k \in \mathcal{N}_i} \frac{\partial\mathcal{L}_i}{\partial f(x_k)} = \frac{f(x_i)}{\tau} \cdot \sum_{k \in \mathcal{N}_i} p_{ik}\ , \ p_{ij} + \sum_{k \in \mathcal{N}_i} p_{ik} = 1 
\end{equation}

Thus, the equation in Theorem 1 is proved:

\begin{equation}
\sum_{k \in \mathcal{N}_i} \frac{\partial \mathcal{L}_i}{\partial f(x_k)} = \frac{1}{\tau} \cdot (1 - p_{ij}) \cdot f(x_i)= -\frac{\partial \mathcal{L}_i}{\partial f(x_j)}
\end{equation}

From Theorem 1, it follows that in existing contrastive loss, the gradient contribution of a single positive sample equals the sum of gradient contributions from all negative samples. This imbalance becomes particularly prominent in datasets with a large number of samples, where the parameter update direction is dominated by positive sample gradients, while the role of negative samples is diluted. This causes nodes to over-focus on themselves and distance themselves from all other nodes, undermining the contrastive channel's ability to capture global information. To address this issue, we propose two strategies for augmenting positive samples. 

After augmenting positive samples, Theorem 1 can be generalized as: for any anchor node $i$, let \(\mathcal{P}_i\) denote the set of positive samples of $i$. Through a proof process analogous to the original theorem, the following equation holds:

\begin{equation}
\sum_{k \in \mathcal{\mathcal{P}}_i} \frac{\partial\mathcal{L}_{i}}{\partial f(x_{k})} = - \sum_{j \in \mathcal{N}_i}\frac{\partial \mathcal{L}_{i}}{\partial f(x_{j})}
\end{equation}

Therefore, augmenting positive samples can alleviate gradient imbalance and enhance the contrastive channel's ability to capture global information.

\paragraph{\textbf{Positive Sample Augmentation Strategy Based on Meta-path Connection Count.}}

For nodes $i$ and $j$, a function $\mathbf{C}_i(j)$ is defined to count the number of meta-paths connecting them:

\begin{equation}
\mathbf{C}_{i}(j)=\sum_{\phi \in \Phi}^{M} \mathbf{1}\left(j \in N_{i}^{\phi_{n}}\right)
\end{equation}

Here, $\mathbf{1}(\cdot)$ is the indicator function. $N_{i}^{\phi_{n}}$ denotes the neighbor set of node $i$ under the meta-path of type $\phi_{n}$. A set $\{S_i = j \mid j \in \mathcal{V} \ and \ \mathbf{C}_i(j) \neq 0 \}$ is constructed and sorted in descending order based on $\mathbf{C}_i(j)$.

A threshold $T_{\text{pos}}$ is defined. If $|S_i| > T_{\text{pos}}$, the top $T_{\text{pos}}$ nodes from $S_i$ are selected as positive samples for node $i$, denoted as $\mathcal{P}_i$. Otherwise, all nodes in $S_i$ are retained. All remaining nodes are considered as negative samples and denoted as $\mathcal{N}_i$.

A positive sample matrix $\mathbf{P}$ is thus obtained, where $\mathbf{P}_{i,j} = 1$ if node $j$ is a positive sample of node $i$.

For any $t \geq 2$, the $t$-hop positive sample matrix $P^{(t)}$ is computed as:
$P^{(t)} = (P^{(t-1)} \cdot P) > 0$. The final combined matrix is obtained by applying element-wise logical OR over the 1-hop to $k$-hop matrices:

\begin{equation}
{P}_{\text{combined}} = \bigvee_{t=1}^{k} {P}^{(t)}
\label{eq:4}
\end{equation}

\({P}_{\text{combined}}\) denotes the positive sample matrix representing fused k hop positive samples. Different values of $k$ are used for different datasets. 

\paragraph{\textbf{Positive Sample Augmentation via Clustering Algorithm.}}

After one round of training, an initial embedding was obtained as $H = g_E(A^{\Phi}, \mathcal{X})$. Given $S$ classes and node representations $H$, a clustering algorithm (K-means) \cite{lloyd1982least} was applied to produce $S$ clusters $\mathcal{C} = \{C_1, C_2, \ldots, C_S\}$, where $C_k$ denotes the index set of samples in the $k$-th cluster, satisfying $\bigcup_{k=1}^{S} C_k = \{1, 2, \ldots, |\mathcal{V}|\}$ and $C_i \cap C_j = \emptyset \quad (i \neq j)$.

For node $i$, $h_i$ is its node representation. Cosine distance $1 - \cos(h_a, h_b)$ was adopted as the distance between node $a$ and $b$. For any node $i \ (i \in C_k)$ in the $k$-th cluster $C_k$, the average distance to nodes in other clusters was defined as:

\begin{equation}
\bar{D}_k(i) = \frac{1}{|\bar{C}_k|} \sum_{m \in \bar{C}_k} \text{dist}(h_i, h_m) \ , \ \bar{C}_k = \{ m \mid m \notin C_k \}
\end{equation}

Then, nodes in $C_k$ were sorted in descending order of $\bar{D}_k(i)$, and the top $K$ nodes were selected as the “key deviated nodes” set $S_k$.

Then, $S_k$ was used to augment the positive sample set for all nodes in the same cluster, defined as:

\begin{equation}
\mathcal{P}(i) = \mathcal{P}(i) \cup S_k \ ,\  \forall i \in C_k
\end{equation}

\paragraph{\textbf{Loss Composition.}}
HetCRF was further trained with these newly augmented heuristic positive samples. The augmented positive sample loss was defined as:

\begin{equation}
\mathcal{L}_{\text{con}} = -\log \frac{\sum_{k \in \mathcal{P}_i} \exp({Z}_{\text{fusion}}^i \cdot {Z}_{\text{schema}}^k / \tau)}{\sum_{j \in \mathcal{P}_i \cup \mathcal{N}_i} \exp({Z}_{\text{fusion}}^j \cdot {Z}_{\text{schema}}^j / \tau)}
\end{equation}

The final objective function \( L \) is defined as a weighted combination of \( L_{\text{feat}} \), \( L_{\text{mp}} \), and \( L_{\text{con}} \).

\begin{equation}
\mathcal{L}_{final} = \lambda_1 \mathcal{L}_{feat} + \lambda_2 \mathcal{L}_{mp} + (1 - \lambda_1 - \lambda_2) \mathcal{L}_{con}
\end{equation}

\section{Experiments}

To comprehensively evaluate the performance of HetCRF, experiments were conducted on four real-world heterogeneous graph datasets: DBLP, Freebase, ACM, and AMiner. Among them, the Aminer and Freebase datasets are categorized into semantically sparse scenarios due to missing node features, while DBLP and ACM are classified into semantically richer scenarios. Ten representative graph representation learning methods were selected as baselines, covering different graph types and learning paradigms. These include the generative self-supervised method for homogeneous graphs (GraphMAE\cite{hou2022graphmae}), supervised methods for heterogeneous graphs (HAN\cite{wang2019heterogeneous}), generative self-supervised methods for heterogeneous graphs (HGMAE\cite{tian2023heterogeneous}, RMR\cite{duan2024reserving}), contrastive self-supervised methods for heterogeneous graphs (HeCo\cite{wang2021self}, MEOW\cite{yu2024heterogeneous}, DiffGraph\cite{li2025diffgraph}, ASHGCL\cite{jiang2025incorporating}, HERO\cite{mo2024self}) and HGMS-C\cite{wang2025homophily}) and generative-contrastive hybrid framework for homogeneous graphs(GCMAE\cite{wang2024generative}). Due to the unavailability of the source code for the hybrid framework GC-HGNN applied to heterogeneous graphs, and the differences in experimental settings reported in the paper, it is not included in our experiments.

\begin{table*}[htbp]
  \centering
  \small
  \renewcommand{\arraystretch}{1.1}
  \setlength{\tabcolsep}{2pt}  
  \begin{tabular}{c|c|c|c c c c c c c c c c|c}
    \toprule
    \textbf{Datasets} & \textbf{Metric} & \textbf{Split}
      & \textbf{\makecell{HeCo \\ KDD'21}} 
      & \textbf{\makecell{GraphMAE \\ KDD'22}} 
      & \textbf{\makecell{HGMAE \\ AAAI'22}} 
      & \textbf{\makecell{MEOW \\ SDM'23}} 
      & \textbf{\makecell{RMR \\ KDD'24}}
      & \textbf{\makecell{GCMAE \\ ICDE'24}}
      & \textbf{\makecell{HERO \\ ICRL'24}} 
      & \textbf{\makecell{DiffGraph \\ WSDM'25}} 
      & \textbf{\makecell{ASHGCL \\ IF'25}}  
      & \textbf{\makecell{HGMS-C \\ SIGIR'25}} 
      & \textbf{\makecell{HetCRF}} \\
    \hline

    \multirow{9}{*}{\centering DBLP} 
      & \multirow{3}{*}{Mi-F1} & 20 
        & 91.97±0.2 & 89.31±0.7 & 91.85±0.5       & 92.64±0.4 & 91.97±0.5 & 90.15±0.4 & 92.07±0.2 & 92.90±0.4          & \uline{93.51±0.4} & \textbf{93.94±0.3} & 93.24±0.4          \\
    \cline{3-14}
      &                       & 40 
        & 90.76±0.3 & 87.80±0.5 & 91.59±0.3       & 91.51±0.3 & 91.15±0.3 & 89.27±0.2 & 91.14±0.3 & 92.01±0.5          & \uline{92.42±0.2} & 91.81±0.1          & \textbf{92.82±0.3} \\
    \cline{3-14}
      &                       & 60 
        & 91.59±0.2 & 89.82±0.4 & 92.50±0.3       & 93.50±0.5 & 92.34±0.2 & 90.11±0.2 & 91.62±0.2 & 92.65±0.5          & \uline{93.68±0.2} & 93.44±0.2          & \textbf{93.69±0.2} \\
    \cline{2-14}
      & \multirow{3}{*}{Ma-F1} & 20 
        & 91.28±0.2 & 87.94±0.7 & 91.22±0.5       & 92.16±0.5 & 91.32±0.4 & 90.88±0.5 & 91.68±0.3 & 92.01±0.2          & \uline{93.05±0.2} & \textbf{93.88±0.2} & 92.80±0.4          \\
    \cline{3-14}
      &                       & 40 
        & 90.34±0.3 & 86.85±0.7 & 91.29±0.3       & 91.15±0.3 & 91.23±0.3 & 90.01±0.2 & 91.33±0.3 & 91.90±0.3          & \uline{92.05±0.1} & 91.21±0.2          & \textbf{92.55±0.3} \\
    \cline{3-14}
      &                       & 60 
        & 90.64±0.3 & 88.07±0.6 & 91.79±0.3       & 93.05±0.5 & 91.60±0.3 & 91.28±0.1 & 91.72±0.2 & 92.13±0.3          & 92.46±0.1       & \uline{93.08±0.1}    & \textbf{93.12±0.2} \\
    \cline{2-14}
      & \multirow{3}{*}{AUC}   & 20 
        & 98.32±0.1 & 92.23±3.0 & 98.71±0.1       & 98.87±0.2 & 98.28±0.1 & 95.01±2.5 & 98.25±0.2 & \uline{98.92±0.5}    & 98.81±0.2       & 98.82±0.1          & \textbf{99.06±0.1} \\
    \cline{3-14}
      &                       & 40 
        & 98.06±0.1 & 91.76±2.5 & 98.48±0.1       & 98.55±0.2 & 98.21±0.1 & 94.99±2.2 & 98.34±0.1 & 98.42±0.1          & 98.42±0.3       & \uline{98.70±0.1}    & \textbf{98.79±0.1} \\
    \cline{3-14}
      &                       & 60 
        & 98.59±0.1 & 91.63±2.5 & 98.92±0.1       & 99.05±0.1 & 98.56±0.1 & 95.23±1.5 & 98.44±0.1 & \textbf{99.15±0.3} & 98.95±0.2       & 99.01±0.1          & \uline{99.06±0.1}    \\
    \hline

    \multirow{9}{*}{\centering Freebase} 
      & \multirow{3}{*}{Mi-F1} & 20 
        & 61.72±0.6 & 64.88±1.8 & 64.45±1.4       & 58.86±0.7 & 61.27±1.0 & 65.12±1.5 & 59.73±1.8 & 64.55±1.9          & \uline{67.01±0.3} & 65.18±1.3          & \textbf{68.68±1.5} \\
    \cline{3-14}
      &                       & 40 
        & 64.03±0.7 & 62.34±1.0 & \uline{66.96±0.8} & 59.64±0.5 & 63.79±1.2 & 64.23±1.2 & 62.10±1.6 & 65.07±1.2          & 66.22±0.5       & 66.30±1.0          & \textbf{70.06±0.8} \\
    \cline{3-14}
      &                       & 60 
        & 63.61±1.6 & 59.48±6.2 & 64.55±2.1       & 59.21±0.4 & 64.18±1.4 & 64.12±1.3 & 62.82±1.5 & 66.32±1.9          & 64.80±0.3       & \uline{66.58±0.8}    & \textbf{70.59±1.0} \\
    \cline{2-14}
      & \multirow{3}{*}{Ma-F1} & 20 
        & 59.23±0.7 & 59.04±1.0 & 61.73±1.2       & 55.01±0.9 & 60.42±1.1 & 60.78±0.7 & 56.47±1.1 & 62.93±1.1          & \uline{63.02±0.5} & 61.60±1.0          & \textbf{64.89±0.9} \\
    \cline{3-14}
      &                       & 40 
        & 61.19±0.6 & 56.40±1.1 & 63.44±0.9       & 55.19±0.8 & 61.51±0.7 & 61.74±0.6 & 57.29±0.8 & \uline{64.87±0.8}    & 62.68±0.3       & 62.74±0.8          & \textbf{67.07±1.2} \\
    \cline{3-14}
      &                       & 60 
        & 60.13±1.3 & 51.73±2.3 & 61.18±1.7       & 57.46±0.5 & 61.26±0.9 & 60.88±0.8 & 56.93±0.6 & \uline{63.21±1.4}    & 60.95±0.6       & 62.46±0.6          & \textbf{66.37±0.9} \\
    \cline{2-14}
      & \multirow{3}{*}{AUC}   & 20 
        & 76.22±0.8 & 72.60±0.2 & 77.48±1.2       & 73.51±0.7 & 75.39±0.8 & 74.89±1.5 & 74.12±1.7 & \uline{77.72±0.9}    & 77.29±0.5       & 77.59±0.9          & \textbf{81.07±1.4} \\
    \cline{3-14}
      &                       & 40 
       & 78.44±0.5 & 72.44±1.6 & 79.13±0.6       & 75.32±0.8 & 73.28±0.5 & 73.28±0.8 & 74.55±1.6 & \uline{79.79±0.8}    & 78.35±0.4       & 79.30±0.8          & \textbf{81.99±0.7} \\
    \cline{3-14}
      &                       & 60 
        & 78.04±0.4 & 70.66±1.6 & 78.03±2.1       & 75.44±0.7 & 77.84±0.6 & 73.12±0.6 & 74.22±1.6 & \uline{79.12±1.3}    & 78.19±0.4       & 78.72±0.6          & \textbf{81.05±0.8} \\
    \hline

\multirow{9}{*}{\centering ACM} 
  & \multirow{3}{*}{Mi-F1} & 20 
    & 88.13±0.8 & 82.48±1.9 & 89.29±0.3       & 90.63±0.3 & 88.94±0.3 & 84.11±0.7 & 88.71±0.8 & 90.42±1.0          & 90.75±0.6       & \uline{92.77±0.4}    & \textbf{93.32±0.2} \\
\cline{3-14}
  &                       & 40 
    & 87.45±0.5 & 82.93±1.1 & 89.40±0.3       & 90.14±0.4 & 88.56±0.4 & 85.54±0.5 & 88.65±0.5 & 90.95±0.7          & 90.92±0.1       & \uline{92.21±0.3}    & \textbf{92.30±0.4} \\
\cline{3-14}
  &                       & 60 
    & 88.71±0.5 & 80.77±1.1 & 89.72±0.7       & 90.76±0.3 & 87.76±0.4 & 83.79±0.3 & 89.62±0.4 & 90.73±0.6          & \uline{92.25±0.1} & 92.19±0.2          & \textbf{93.28±0.2} \\
\cline{2-14}
  & \multirow{3}{*}{Ma-F1} & 20 
    & 88.56±0.8 & 82.26±1.5 & 89.59±0.3       & 90.62±0.5 & 89.02±0.3 & 85.65±1.5 & 88.37±1.3 & 90.81±0.9          & 91.43±0.2       & \uline{93.07±0.4}    & \textbf{93.43±0.2} \\
\cline{3-14}
  &                       & 40 
    & 87.61±0.5 & 82.00±1.1 & 89.53±0.3       & 89.85±0.4 & 88.67±0.4 & 83.12±0.9 & 88.59±1.1 & 89.96±0.9          & 91.25±0.1       & \textbf{92.38±0.2} & \uline{92.35±0.3}    \\
\cline{3-14}
  &                       & 60 
    & 89.04±0.5 & 80.29±1.0 & 89.85±0.7       & 90.84±0.3 & 87.96±0.4 & 84.23±0.5 & 89.32±0.6 & 91.73±0.7          & \uline{92.46±0.1} & 92.42±0.4          & \textbf{93.39±0.2} \\
\cline{2-14}
  & \multirow{3}{*}{AUC}   & 20 
    & 96.49±0.3 & 92.09±0.5 & 97.69±0.1       & 97.60±0.2 & 97.09±0.1 & 93.66±0.3 & 96.48±0.4 & 97.56±0.3          & 97.86±0.2       & \uline{98.39±0.4}    & \textbf{98.71±0.1} \\
\cline{3-14}
  &                       & 40 
    & 96.40±0.4 & 92.65±0.5 & 97.52±0.1       & 97.74±0.1 & 96.75±0.1 & 91.11±0.5 & 96.50±0.3 & 97.72±0.3          & 98.24±0.3       & \uline{98.45±0.2}    & \textbf{98.84±0.1} \\
\cline{3-14}
  &                       & 60 
    & 96.55±0.3 & 91.49±0.6 & 97.87±0.1       & 98.05±0.1 & 95.38±0.6 & 91.32±0.7 & 97.03±0.2 & 97.66±0.2          & 98.08±0.2       & \uline{98.46±0.2}    & \textbf{98.55±0.1} \\
\hline

\multirow{9}{*}{\centering Aminer} 
  & \multirow{3}{*}{Mi-F1} & 20 
    & 78.81±1.3 & 68.21±0.3 & 78.93±0.9       & 76.26±0.5 & 78.96±1.1 & 70.21±0.5 & 80.68±1.3 & 79.99±0.7          & 80.40±0.6       & \uline{82.61±1.3}    & \textbf{84.94±0.8} \\
\cline{3-14}
  &                       & 40 
    & 80.53±0.7 & 74.23±0.2 & 82.40±0.9       & 75.90±0.6 & 80.85±0.9 & 76.34±0.8 & 82.24±0.9 & 81.36±1.1          & 82.07±0.9       & \uline{84.12±0.9}    & \textbf{85.80±0.6} \\
\cline{3-14}
  &                       & 60 
    & 82.46±1.4 & 72.28±0.2 & 81.66±0.6       & 77.92±0.4 & 82.17±0.7 & 75.35±0.2 & 82.35±0.6 & 81.10±1.2          & 81.65±0.7       & \uline{85.13±0.6}    & \textbf{85.81±0.4} \\
\cline{2-14}
  & \multirow{3}{*}{Ma-F1} & 20 
    & 71.38±1.1 & 62.64±0.2 & 70.33±1.0       & 69.79±0.9 & 72.41±0.7 & 65.24±0.3 & 72.79±1.3 & 71.40±0.9          & 72.80±1.8       & \uline{74.88±1.0}    & \textbf{77.96±1.0} \\
\cline{3-14}
  &                       & 40 
    & 73.75±0.5 & 68.17±0.2 & 74.58±1.0       & 70.26±1.0 & 72.41±0.7 & 70.15±0.5 & 75.62±0.8 & 75.37±1.0          & 75.42±1.1       & \uline{76.93±0.6}    & \textbf{79.68±0.7} \\
\cline{3-14}
  &                       & 60 
    & 75.80±1.8 & 68.21±0.2 & 74.61±0.6       & 72.79±1.0 & 74.72±0.8 & 69.12±1.5 & 75.17±0.6 & 75.32±0.6          & 75.58±1.2       & \uline{78.52±0.5}    & \textbf{80.39±0.5} \\
\cline{2-14}
  & \multirow{3}{*}{AUC}   & 20 
    & 90.82±0.6 & 86.29±4.1 & 92.22±0.5       & 90.45±0.2 & 75.36±0.5 & 88.89±1.4 & 90.94±0.4 & 90.12±1.1          & 90.55±0.5       & \uline{92.96±0.3}    & \textbf{94.47±0.4} \\
\cline{3-14}
  &                       & 40 
    & 92.11±0.6 & 89.98±0.0 & {\ul 94.28±0.3} & 92.43±0.3 & 91.38±0.9 & 90.12±0.8 & 91.22±0.3 & \uline{94.34±0.4}    & 93.75±0.4       & 93.45±0.4          & \textbf{94.90±0.2} \\
\cline{3-14}
  &                       & 60 
    & 92.40±0.7 & 88.32±0.0 & 93.78±0.5       & 92.64±0.2 & 92.14±0.4 & 90.45±0.7 & 92.14±0.2 & \uline{94.12±0.7}    & 92.70±0.5       & 93.79±0.3          & \textbf{95.77±0.3} \\
    \bottomrule
  \end{tabular}
  \caption{Node classification performance comparison across datasets under various label splits. Bold indicates the best; underlined indicates the second best.(As the source code is inaccessible, the HGMS-C model's results are referenced from its original paper. IF'25 means Information Fusion'25.)}
  \label{tab:full-ablation}
\end{table*}

\subsection{Node Classification}

We concatenate the embeddings learned by the HAN encoder and GCN to train a linear classifier for model evaluation.\textbf{Table \ref{tab:full-ablation}} presents the node classification results across all models. HetCRF achieved the best performance in most cases. 

In particular, on the AMiner and Freebase datasets with a label rate of 40\%, the Macro-F1 score of HetCRF increased by 2.75\% and 2.2\%, respectively, compared to the state of the suboptimal baseline. Experiments demonstrate that HetCRF significantly outperforms other models in feature-missing dataset scenarios. 

\subsection{Node clustering}

We further perform K-means clustering to verify the quality
of learned node embeddings. We adopt Normalized Mutual Information (NMI) and Adjusted Rand Index (ARI) as evaluation metrics. \textbf{Table \ref{tab:nmi_ari_all}} presents the clustering results of all models, with HetCRF
demonstrating superior performance over the existing baselines on the four datasets.

\begin{table}[H]
\centering
\small
\renewcommand{\arraystretch}{1.0}
\setlength{\tabcolsep}{3pt}
\begin{tabular}{c|cc|cc|cc|cc}
\toprule
\multirow{2}{*}{Method} & \multicolumn{2}{c|}{ACM} & \multicolumn{2}{c|}{DBLP} & \multicolumn{2}{c|}{AMiner} & \multicolumn{2}{c}{Freebase} \\
\cline{2-9}
 & NMI & ARI & NMI & ARI & NMI & ARI & NMI & ARI \\
\hline
DGI       & 51.73          & 41.16          & 59.23          & 61.85          & 22.06          & 15.93          & 18.34          & 11.29          \\
DMGI      & 51.66          & 46.64          & 70.06          & 75.46          & 19.24          & 20.09          & 16.38          & 16.91          \\
HeCo      & 59.26          & 58.36          & 71.58          & 76.93          & 29.02          & 18.72          & 18.25          & 18.42          \\
HGMAE     & 64.57          & 67.85          & 71.32          & 77.43          & 37.70          & 34.38          & 16.46          & 15.80          \\
MEOW      & 51.12          & 45.20          & 73.78          & 76.58          & 24.05          & 17.66          & 13.25          & 14.40          \\
RMR       & 54.37          & 45.65          & 73.22          & 77.67          & 30.22          & 31.59          & 18.33          & 18.15          \\
HERO      & 58.41          & 57.89          & 71.12          & 75.01          & 30.11          & 31.48          & 15.34          & 15.56          \\
DiffGraph & 63.69          & 62.11          & 74.01          & 77.63          & 29.54          & 37.95          & 15.54          & 17.98          \\
ASHGCL    & 63.76          & 61.82          & {\ul 76.29}    & {\ul 81.67}    & 28.06          & 38.22          & 17.35          & 18.02          \\
HGMS-C    & \textbf{71.97} & \uline{74.18}    & -              & -              & \uline{46.56}    &  \uline{50.28}    & \uline{22.28}    & \uline{23.14}    \\
HetCRF    & \uline{69.48}    & \textbf{75.07} & \textbf{76.87} & \textbf{82.16} & \textbf{48.82} & \textbf{53.12} & \textbf{22.52} & \textbf{23.45} \\
\bottomrule
\end{tabular}
\caption{Node clustering performance (NMI and ARI) comparison on four datasets (As the source code is inaccessible, the HGMS-C model's results are referenced from its original paper.)}
\label{tab:nmi_ari_all}
\end{table}

\begin{figure}[H]
    \centering
    \includegraphics[width=0.5\textwidth]{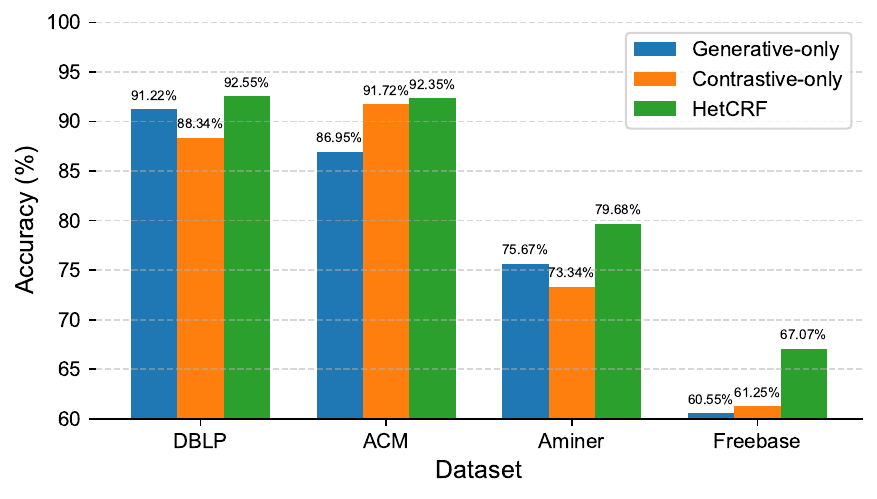}
    \caption{Ablation results comparing different training strategies(Macro-F1,40\%)}
    \label{fig:bar_chart}
    \Description{}
\end{figure}

\begin{figure}[htbp]
    \centering
    \includegraphics[width=0.45\textwidth]{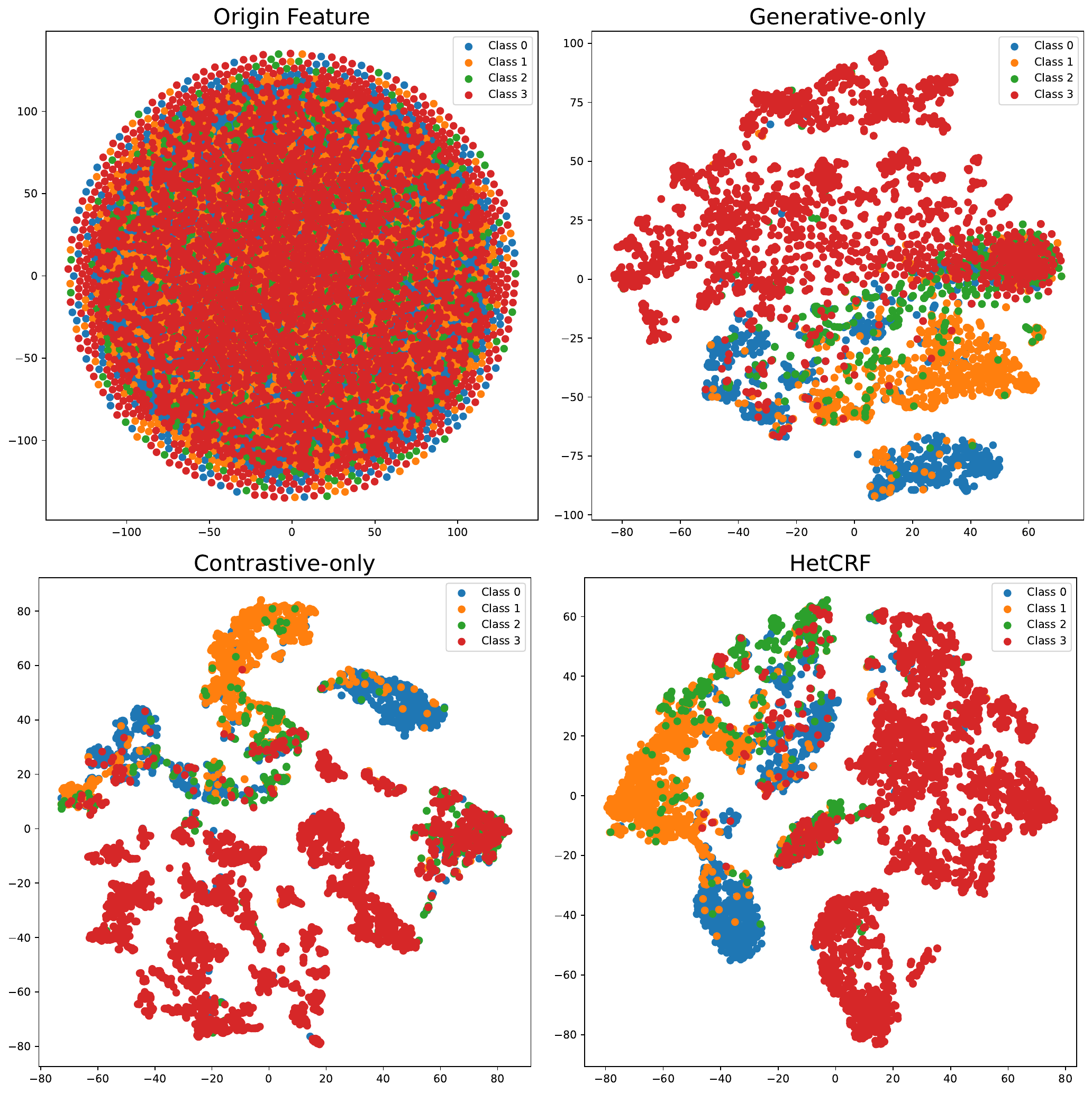}
    \caption{Comparison of t-SNE visualization effects using different methods (AMiner)}
    \label{fig:merged_SNE1}
    \Description{}
\end{figure}

\subsection{Ablation Study}

\subsubsection{Validation of the Hybrid Framework's Effectiveness}

\paragraph{\textbf{A}} As shown in \textbf{Figure \ref{fig:bar_chart}}, we evaluated the performance of three different training strategies under a 20\% label rate on four benchmark datasets: (1) generative-only method; (2) contrastive-only method; (3) hybrid framework combining generative and contrastive approaches(HetCRF). Macro-F1 score was used as the evaluation metric. The experimental results clearly demonstrate the superiority of the dual-channel strategy. To further strengthen this evidence, we present a t-SNE visualization in \textbf{Figure \ref{fig:merged_SNE1}}. The experimental results validate our theoretical analysis: the contrastive channel is capable of better capturing global information and reducing embedding entanglement in the generative channel. Meanwhile, the local information capture capability of the generative channel prevents the embeddings in the contrastive channel from being dispersed.

\subsubsection{Validation of Augmented Positive Samples' Effectiveness}

\paragraph{\textbf{B}} To explore the effectiveness of two positive sample augmentation strategies, We applied the four strategies of no positive sample augmentation(Self-only), meta-path connection counting-based positive sample augmentation only(w/o\_Cluster), clustering-based positive sample augmentation only( w/o\_MPC), and the full model(HetCRF) to the node classification task on each of the four benchmark datasets and denote the results of the Macro-F1 metrics under the 40\% label rate scenario in \textbf{Table \ref{tab:clustering_nmi}}. Experiments show that positive sample augmentation can effectively improve the representation ability of the encoder.

\begin{table}[htbp]
\centering
\small
\begin{tabular}{cccccc}
\toprule
Dataset & Self-only & w/o\_Cluster & w/o\_MPC & HetCRF \\
\midrule
dblp      & 92.04 & \textbf{92.75} & 90.55 & \uline{92.55} \\
acm       & 92.06 & 91.24 & \textbf{92.51} & \uline{92.35} \\
freebase  & 60.06 & 64.06 & \uline{66.65} & \textbf{67.07} \\
aminer    & 74.12 & \uline{78.10} & 77.23 & \textbf{79.68} \\
\bottomrule
\end{tabular}
\caption{Evaluation of positive sample augmentation strategies (Macro-F1, 40\%)}
\label{tab:clustering_nmi}
\end{table}

\begin{figure}[htbp]
    \centering
    \includegraphics[width=0.45\textwidth]{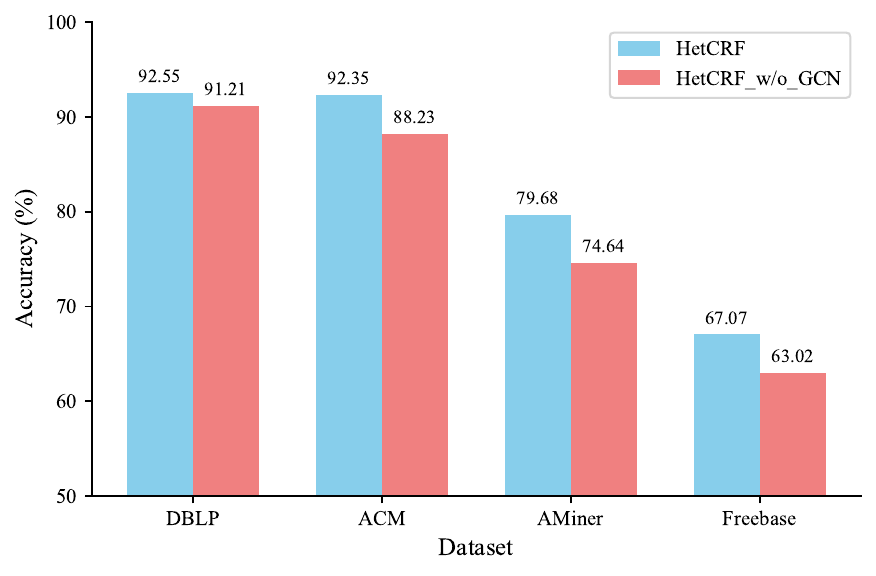}
    \caption{Effect of node classification with or without view building and GCN aggregation procedures on different datasets(Macro-F1,40\%)}
    \label{fig:aug_bar_chart}
    \Description{}
\end{figure}

\begin{figure}[htbp]
    \centering
    \includegraphics[width=0.45\textwidth]{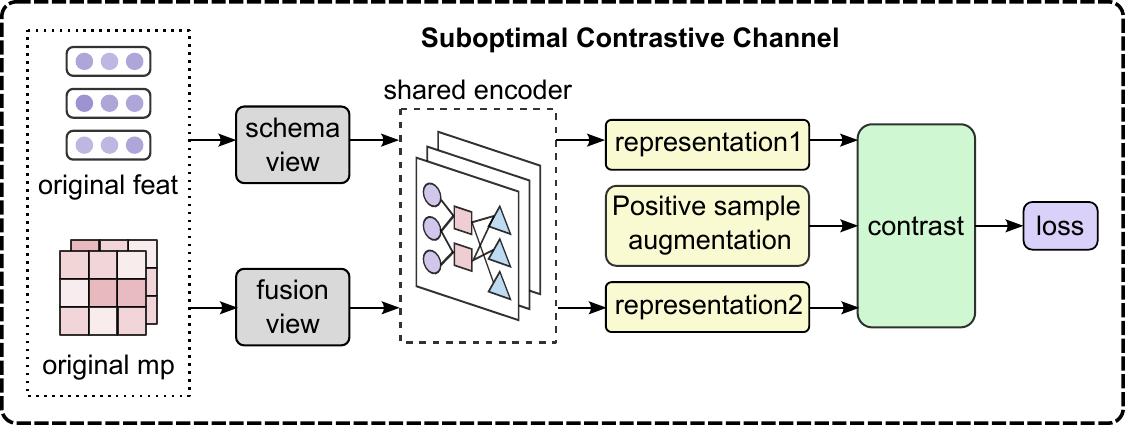}
    \caption{The contrastive learning channel of \(HetCRF_{w/o-_{GCN}}\)}
    \label{fig:con_model}
    \Description{}
\end{figure}

\subsubsection{Validation of Enhanced Contrastive View Effectiveness}

\paragraph{\textbf{C}} Compared with existing contrastive frameworks that obtain enhanced contrastive views from raw data, we derive contrastive views through the "View Construction and GCN Aggregation" procedure in \textbf{Figure \ref{fig:model}}. To validate the effectiveness of the procedure, we propose the model $HetCRF_{w/o-_{GCN}}$. \(HetCRF_{w/o-_{GCN}}\) removes the "View Construction and GCN Aggregation" procedure from the model and constructs contrastive views via augmentation on the original graph, with the detailed architecture illustrated in \textbf{Figure\ref{fig:con_model}}.

We applied HetCRF and $HetCRF_{w/o-_{GCN}}$ on four benchmark datasets and presented the model accuracy results as bar charts in \textbf{Figure \ref{fig:aug_bar_chart}}. We found that this procedure is effective for all datasets. Notably, for datasets with missing initial features (e.g., Aminer, Freebase), the procedure achieves improvements of 5.04\% and 4.05\%, respectively. This result validates our analysis: HetCRF’s newly proposed contrastive channel can adapt to scenarios with sparse semantics.

\subsubsection{Experimental Validation of Generalizability}

\paragraph{\textbf{D}} To validate the generalization performance of HetCRF in semantically sparse scenarios, we transformed the DBLP and ACM datasets into those with missing node features by removing their original features. We conducted node classification experiments on HetCRF and four state-of-the-art baselines: RMR\cite{duan2024reserving}, HERO\cite{mo2024self}, DiffGraph\cite{li2025diffgraph} and ASHGCL\cite{jiang2025incorporating}. (Due to the unavailability of the source code for HGMS-C\cite{wang2025homophily}, we did not conduct experiments on HGMS-C) \textbf{Table \ref{tab:featureless_results}} details their Macro-F1 performance at a 40\% label rate, with HetCRF showing significant improvements.

This experiment demonstrates the generalizability of the HetCRF framework on datasets with missing node features.

\begin{table}[htbp]
\centering
\small
\setlength{\tabcolsep}{3pt}
\begin{tabular}{@{}cccccc@{}}
\toprule
Dataset & RMR & HERO & DiffGraph & ASHGCL & HetCRF \\
\midrule
\texttt{dblp\_w/o feats} & \uline{91.12} & 90.31 & 90.23 & 89.12 & \textbf{92.12} \\
\texttt{acm\_w/o feats}  &81.23 & 80.54 & \uline{82.31} & 79.66 & \textbf{83.09} \\
\bottomrule
\end{tabular}
\caption{Node classification performance comparison after removing initial features(Macro-F1,40\%)}
\label{tab:featureless_results}
\end{table}

\subsubsection{Hyperparameter Experiments}

\begin{figure}[htbp]
    \centering
    \includegraphics[width=0.45\textwidth]{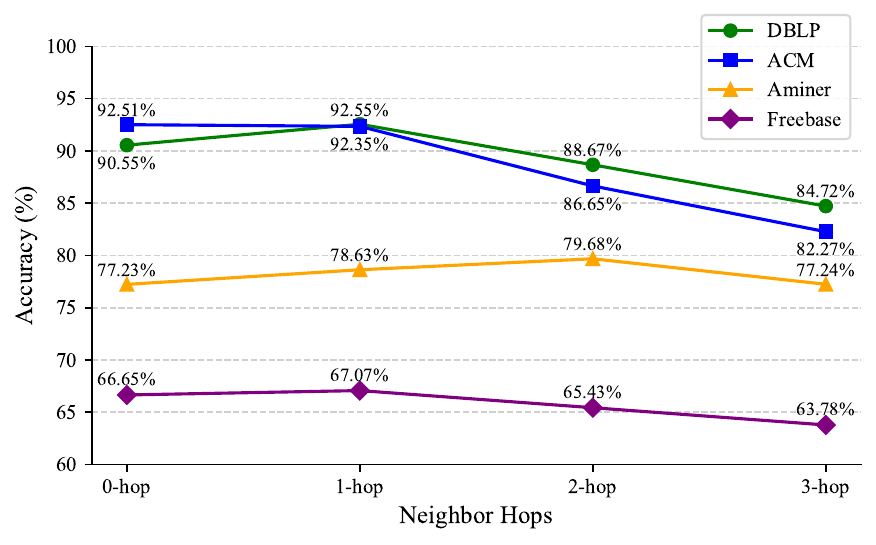}
    \caption{Impact of positive sample augmentation range on model performance (Macro-F1,40\%)}
    \label{fig:line_chart}
    \Description{}
\end{figure}

\paragraph{\textbf{E}} To investigate the impact of the hyperparameter k in Equation(\ref{eq:4}) on model performance, we adjusted the value of k in the meta-path-based positive sample augmentation strategy to control the number of hops (k-hop) for augmenting positive samples, while keeping the clustering-based augmentation strategy unchanged. Comparative experiments were conducted on four benchmark datasets. We plotted a comparison line chart of the model accuracy under different augmentation strategies (\textbf{Figure \ref{fig:line_chart}}) to visually demonstrate the effect of positive sample augmentation on model performance.

\begin{figure}[htbp]
    \centering
    \includegraphics[width=0.45\textwidth]{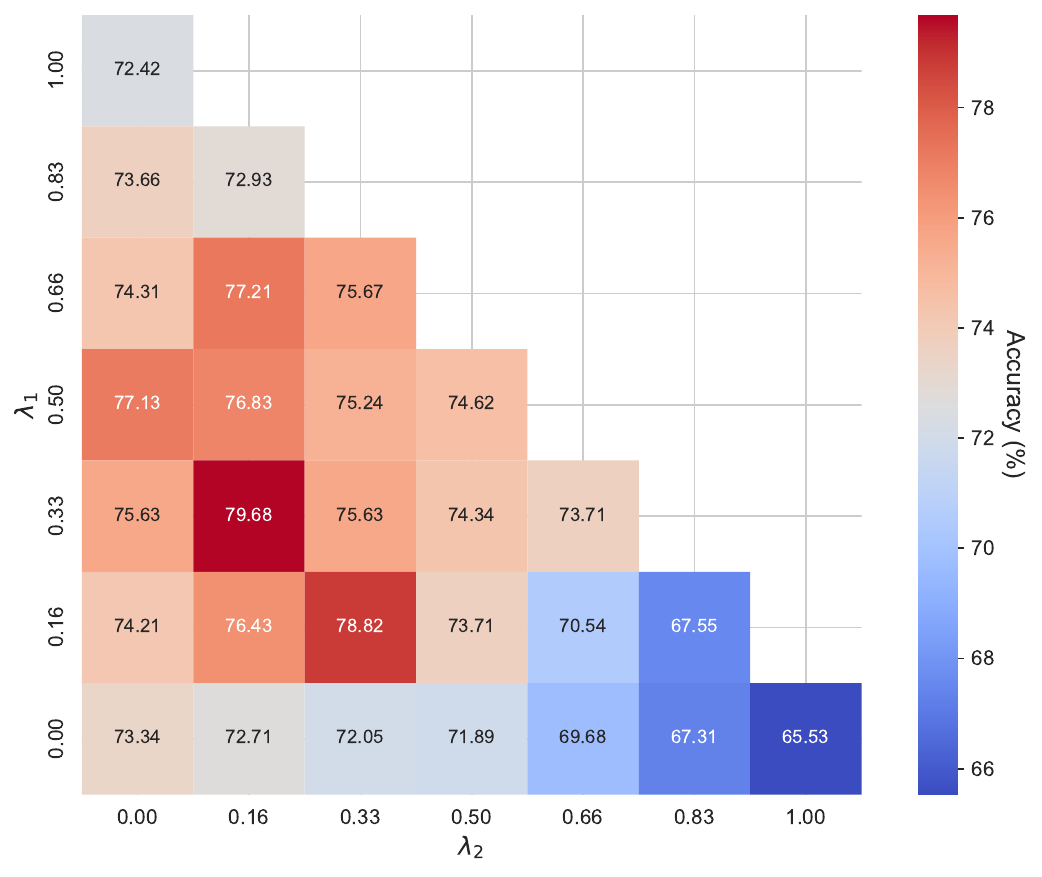}
    \caption{Heatmap of loss weight adjustment impact on model performance(Aminer,Macro-F1,40\%)}
    \label{fig:heat_map}
    \Description{}
\end{figure}

\paragraph{\textbf{F}} To explore the relative contributions of generative and contrastive learning to overall model performance, we adjusted the loss weight parameter loss weight and visualized the accuracy changes under different weight settings using a heatmap, as shown in \textbf{Figure \ref{fig:heat_map}}.

\section{Conclusions}

In this paper, we identify and formalize the challenges faced by existing semi-supervised learning (SSL) hybrid frameworks on heterogeneous graphs. To address these issues, we propose a novel contrastive-reconstruction framework, HetCRF. By introducing an innovative augmented view generation strategy, HetCRF significantly enhances the expressiveness of the contrastive channel under semantically sparse conditions. Moreover, we theoretically demonstrate the gradient imbalance between positive and negative samples present in existing methods, and propose two positive sample augmentation strategies to mitigate this issue. The effectiveness of HetCRF is supported by both theoretical analysis and empirical results. Across four heterogeneous graph datasets, HetCRF consistently outperforms nearly all baseline models. Notably, on datasets with missing features such as Aminer, HetCRF improves the Micro-F1 score by 3.08\% under a 20\% label rate compared to the previous best model.

Furthermore, HetCRF demonstrates strong generalization capabilities in semantically sparse scenarios (e.g., missing initial node features), providing valuable insights for the development of contrastive learning methods applied to semantically sparse scenarios in the future.



\section*{Generative AI Disclosure Statement}
In this paper, generative AI tools (such as ChatGPT) were used solely for proofreading and language polishing of English expressions. All academic viewpoints, experimental data, methodologies, tables, figures and conclusions originate from our original work. We bear full responsibility for the accuracy, integrity, and originality of all scholarly content in this work.

\bibliographystyle{ACM-Reference-Format}
\bibliography{sample-base}


\end{document}